\documentclass[parskip=half,abstracton]{scrartcl}

\usepackage[utf8]{inputenc}
\usepackage{graphicx}
\usepackage[round]{natbib}
\usepackage{authblk}
\usepackage[dvipsnames]{xcolor}
\usepackage{hyperref}
\hypersetup{colorlinks=true,linkcolor=MidnightBlue,citecolor=MidnightBlue,urlcolor=MidnightBlue}

\title{A History of Metaheuristics\thanks{To appear in R.~Martí, P.~Pardalos, and M.~Resende, Handbook of Heuristics, Springer.}}
\author{Kenneth Sörensen \and Marc Sevaux \and Fred Glover}
\date{}

\begin{document}

\maketitle

\begin{abstract} 
    \noindent This chapter describes the history of metaheuristics in five distinct periods, starting long before the first use of the term and ending a long time in the future.
\end{abstract}

\section{Introduction}


Even though people have used heuristics throughout history, and the human brain is equipped with a formidable heuristic engine to solve an enormous array of challenging optimization problems, the \emph{scientific study} of heuristics (and, by extension, metaheuristics)  is a relatively young endeavour. It is not an exaggeration to claim that  the field of (meta)heuristics, especially compared to other fields of study like physics, chemistry, and mathematics, has yet to reach a mature state. Nevertheless, enormous progress has been made since the first metaheuristics concepts were established. In this chapter, we will attempt to describe the historical developments this field of study has gone through since its earliest days.  

\begin{figure}[htbp]
    \centering
    \includegraphics[width=6cm]{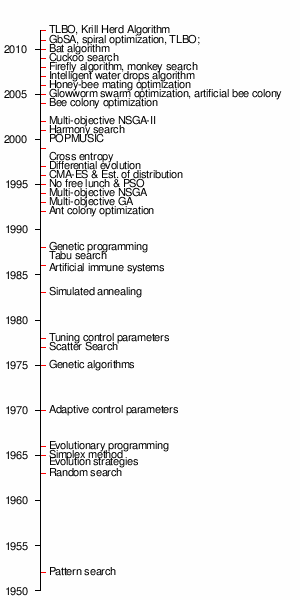}
    \caption{``Most important contributions'' list as it appeared on Wikipedia until April 8, 2013}
    \label{fig:wikipedia}
\end{figure}

No history is ever neutral, and a \textcolor{black}{history of metaheuristics} ---~or any other topic~--- can be written in many different ways.  A straightforward (one could say ``easy'') history of metaheuristics would consist of an annotated and chronological list of metaheuristic methods. Useful as such a list may be, it suffers from a lack of insight into the development of the field as a whole. To illustrate this viewpoint, consider the list in Figure~\ref{fig:wikipedia}, that appeared on Wikipedia until April 8, 2013 to illustrate the ``most important contributions'' in the field of metaheuristics. It is our opinion that such a list is not particularly enlightening (and neither was the article that contained it) when it comes to explaining the evolution of the field of metaheuristics.

Taking a bird's eye view of the field of metaheuristics, one has to conclude that there has been a large amount of progressive insight over the years. Moreover, this progressive insight has not reached its end point: the way researchers and practitioners look at metaheuristics is still continually shifting. Even the answer to the question what a metaheuristic is, has changed quite a lot since the word was first coined in the second half of the 1980s, as will become clear in the rest of this chapter. In our view, it is this shifting viewpoint that deserves to be written down, as it allows us to truly understand the past and perhaps learn a few lessons that could be useful for the future development of research in metaheuristics. We did not limit our discussion in this chapter to metaheuristics that have been formally written down and published. When studying the history of metaheuristics with an open mind, one has to conclude that people have been using heuristics and metaheuristics long before the term even existed.  

We have therefore adopted a different approach to write ``our'' history of metaheuristics. Our approach starts well before the term ``metaheuristic'' was coined and is based on the premise that people have looked at metaheuristics through different sets of glasses over the years. The way in which people ---~not only researchers~--- have interpreted the different metaheuristic concepts has shaped the way in which the field has been developing. To understand the design choices that people have been making when developing metaheuristic optimization algorithms, it is paramount that these choices are understood in relationship to the trends and viewpoints of the time during which the development took place.

Our history divides time in five distinct \textcolor{black}{periods}. The crispness of the boundaries between each pair of consecutive periods, however, is a gross simplification of reality. The real (if one can use that word) time periods during which the paradigm shifts took place are usually spaced out over several years, but it is difficult, if not impossible, to trace the exact moments in time at which the paradigm shifts began and ended. More importantly, not every researcher necessarily makes the transition at the same time.  
\begin{itemize}
    \item The \emph{pre-theoretical} period (until c.~1940), during which heuristics and even metaheuristics are used but not formally studied.
    \item The \emph{early} period (c.~1940 -- c.~1980), during which the first formal studies on heuristics appear. 
    \item The \emph{method-centric} period (c.~1980 -- c.~2000), during which the field of metaheuristics truly takes off and many different methods are proposed.
    \item The \emph{framework-centric} period (c.~2000 -- now), during which the insight grows that metaheuristics are more usefully described as frameworks, and not as methods.
    \item The \emph{scientific} period (the future), during which the design of metaheuristics becomes a science instead of an art.
\end{itemize}

Until recently, a clear definition of the word metaheuristic has been lacking,  and it could be argued that it is still disputed. In this chapter, we adopt the definition of \citet{sorensen.glover:13}.

\begin{quote}
``A metaheuristic is a high-level problem-independent algorithmic framework that provides a set of guidelines or strategies to develop heuristic optimization algorithms. The term is also used to refer to a problem-specific implementation of a heuristic optimization algorithm according to the guidelines expressed in such a framework.''
\end{quote}

The term ``metaheuristic'' has been used (and is used) for two entirely different things. One is a high-level framework, a set of concepts and strategies that blend together, and offer a perspective on the development of optimization algorithms. In this sense, variable neighborhood search \citep{mladenovic.hansen:97} is nothing more (or less) than the idea to use different local search operators to work on a single solution, together with a perturbation operator once all neighborhoods have reached a local optimum. There is a compelling motivation, as well a large amount of empirical evidence, as to why multi-neighborhood search is indeed a very good idea. This motivation essentially comes down to the fact that a local optimum for one local search operator (or one neighborhood structure) is usually not a local optimum for another local search operator. The idea to switch to a different local search operator once a local optimum has been found, is therefore both sensible and in practice extremely powerful.  

The second meaning of the term ``metaheuristic'' denotes a specific implementation of an algorithm based on such a framework (or on a combination of concepts from different frameworks) designed to find a solution to a specific optimization problem. The variable neighborhood search (-based) algorithm for the location-routing problem by \citet{jarboui.derbel.ea:13} is an example of a metaheuristic in this sense.

In this chapter, we will use the term ``metaheuristic framework'' to refer to the first sense and ``metaheuristic algorithm'' to refer to the second sense of the word ``metaheuristic''.

As mentioned, a history of any topic is not neutral. We therefore do not attempt to hide the fact that certain ways in which the field has been progressing seem to us less useful, and sometimes even harmful to the development of the field in general. For example, many of the entries that appear on the list in Figure~\ref{fig:wikipedia} are, in our view, not ``important contributions'' at all, but rather marginal additions to a list of generally useless ``novel'' metaphor-based methods that are best forgotten as quickly as possible.

\section{Period 0: The \textcolor{black}{pre-theoretical period}}

Optimization problems are all around us. When we decide upon the road to take to work, when we put the groceries in the fridge, when we decide which investments to make so as to maximize our expected profit, we are essentially solving an optimization problem (a shortest path problem, a packing problem, and a knapsack problem respectively). For human beings (and many animal species), solving an optimization problem does not require any formal training, something which is immediately clear from the examples given here. The difference between exact solutions and approximate solutions, the difference between easy and hard optimization problems or between fast (polynomial) and slow (exponential) algorithms are all moot to the average problem-solver. 

Indeed, the human mind seems to be formidably equipped from early childhood on to solve an incredible range of problems, many of which could be easily modelled as optimization problems. Most likely, the ability to solve optimization problems adequately and quickly is one of the most important determinants of the probability of survival in all sentient species, and has therefore been favored by evolution throughout time.
Clearly, the human (and animal) mind solves optimization problems \emph{heuristically} and not exactly, i.e., the solutions produced by the brain are by no means guaranteed to be optimal. Given what we now know about exact solution procedures, this makes perfect sense. When determining the trajectory of a spear to hit a mammoth, it is much more important that this trajectory be calculated quickly rather than optimally. Given our knowledge on exact solution methods, we can now say that the calculation of the exact solution (let us say the solution that has the highest probability to hit the mammoth exactly between the eyes given its current trajectory, the terrain in front of it, its anticipated trajectory changes, the current wind direction, etc.) would almost certainly be found only \emph{after} our target has disappeared on the horizon. Moreover, it would almost certainly require too much computing power from the brain, quickly depleting the body's scarce energy resources. 

Given the diversity of problems the human mind must solve, including problems with which it has no prior experience, there is very little doubt that the human mind has the capacity (whether evolved or learned) to use meta-heuristic strategies. Just like the metaheuristics for optimization that form the subject of this book, such strategies are not heuristics in themselves, but are used to derive heuristics from. For example, when confronted with a new problem to which a solution is not immediately obvious (e.g., determining the trajectory of a spear to hit a mammoth), the human mind will automatically attempt to find similar problems it has solved in the past (e.g., determining the trajectory of a stone to hit a bear) and attempt to derive the rules it has learned by solving this problem. This strategy is called \emph{learning by analogy} \citep{carbonell:83}. Another example is called means-end-analysis \citep{simon:96} and can be summarized as follows: given a current state and a goal state, choose an action that will lead to a new state that is closer to the goal state than the current state. This rule is iteratively applied until the goal state has been reached or no other state can be found closer to the goal state than the current state. Obviously, this strategy is a more general counterpart of all formal optimization heuristics that can be categorized as local search, in which a solution is iteratively improved using small, incremental operations we have come to call \emph{moves}. The technique of \emph{path relinking} \citep{glover:97}, in which an incumbent solution is transformed, one move at a time, into a guiding solution, is another example of a formalized means-end-analysis strategy.

Whereas heuristics (and even metaheuristics) are completely natural to us humans, exact methods seem to be a very recent invention, coinciding with the introduction of the field of Operations Research around WWII. 
On the other hand, even though heuristics have been applied since the first life on earth evolved, the \emph{scientific study} of heuristics also had to wait until the 20th century. It could be hypothesized that heuristics are so natural to us, that we had to wait until a formal theory of optimization, especially of linear programming, had to be developed before anyone considered it a topic worthy of study.

\section{Period 1: The \textcolor{black}{early period}}

In 1945, immediately after WWII, the Hungarian mathematician George P\'olya, then working at Stanford University, published a small volume called ``How to Solve it'' \citep{polya:14}. In his book, he argued that problems can be solved by a limited set of generally applicable strategies, most of which serve to make the problem simpler to solve. The book's focus was not on optimization problems, but on the more general class of ``mathematical'' problems, i.e., problems that can be modelled and solved by mathematical techniques. Nevertheless, most of the solution strategies proposed in P\'olya's book are equally applicable to develop optimization algorithms.

The ``analogy'' principle, e.g., tells the problem-solver to look for another problem that closely resembles the problem at hand, and to which a solution method is known. By studying the similarities and differences between both problems, ideas can be garnered to solve the original problem.  The principle of ``induction'' to solve a problem by deriving a generalization from some examples. The ``auxiliary problem'' idea asks whether a subproblem exists that can help to solve the overall problem. 

Even though it is a bit of stretch to call these principles ``meta-heuristics'', it is clear that the start of the field of OR also marks the age during which people started thinking about more general principles that are useful in the design of heuristic algorithms (or solution methods for other types of problems). A case can be made for the fact that many of P\'olya's principles are still heavily used today by heuristic designers. Looking for similar problems in the literature or elsewhere, and modifying the best-known methods for them to suit the problem at hand (analogy), is an extremely common strategy to arrive at a good heuristic fast. Solving some simple examples by hand, and using the lessons learned from your own (or someone else's) perceived strategy to derive an intelligent solution strategy from (induction), is also a useful technique. Finally, decomposing a problem into smaller subproblems and developing specialized techniques for each of them (auxiliary problem) has proven to be a powerful heuristic design strategy on a large number of occasions. 

What is important is that none of P\'olya's strategies actually solve any problem, nor can they be called ``algorithms'' in themselves. Instead, they are high-level, meta-strategies that are useful to influence the way a heuristic designer thinks about a problem. In that sense at the very least, they are very like the more advanced and specialized metaheuristic frameworks that we have today.

Several very high-level algorithmic ideas also came about around this period. The fact that good solutions can be reached by a constructive procedure, for example, is one of them. A constructive algorithm is one that starts from an empty solution and iteratively adds one element at a time until a complete solution has been formed. Simple rules for selecting this element from the set of all potential elements have led to different types of algorithms. The \emph{greedy} selection rule selects the best (value for each) element at each iteration. Kruskal's or Prim's algorithm for the minimum spanning tree problem, Dijkstra's algorithm for the shortest path problem, etc., are all examples of greedy heuristics \citep{cormen.leiserson.ea:09}.  \emph{Regret} algorithms present a similar class of optimization procedures, that select, at each iteration, the element for which \emph{not} selecting its best value results in the highest penalty cost. Vogel's approximation method \citep{shore:70} for the transportation problem is a well-known example. Again, calling the greedy idea or the regret idea ``metaheuristics''  is a bit of a stretch, but they \emph{are} high-level strategies, and they are \emph{not} algorithms themselves.  

Also during this period, \citet{simon.newell:58} see heuristics specifically as fit to solve what they call ``ill-structured'' problems. Contrary to well-structured problems, such problems cannot be formulated explicitly or solved by known and feasible computational techniques. Their predictions in 1958 have turned out to be slightly optimistic, but it cannot be denied that heuristics have turned out to be more flexible problem-solving strategies than exact methods.  

Even though the heuristics developed in the early period were very simple, the realization that high-level strategies existed that could be used as the basis for the development of heuristics for \emph{any} optimization problem led to insights that paved the way for more complex meta-strategies. Together with the widespread availability of computers, these developments took the field of heuristics into the next period in this history, the method-centric period.

\section{Period 2: The \textcolor{black}{method-centric period}}
\label{sec:method}

Even though the frameworks and ideas developed during what we have called the early period lacked the comprehensiveness of the later developed metaheuristic frameworks like tabu search \citep{glover:86}, it is not too far-fetched to call them early metaheuristics. Like later metaheuristic frameworks, these methods offered ---~in the form of some generally applicable strategies~--- inspiration for the development of optimization algorithms. Of course, these principles still needed to be instantiated for each different optimization problem, but at least the process of coming up with an optimization strategy did not have to start from scratch. 

Much of the work done in the early period can be characterized under the umbrella term of \emph{artificial intelligence} because it involves mimicking human problem-solving behavior and learning lessons from this behavior on a more abstract level.  Starting in the 1960's however, an entirely different line of research into problem-solving methods came to life. These methods used an analogy with life's main problem-solving method: evolution. 

Evolution by natural selection has been called ``the best idea ever'' \citep{chu:14}. No single idea explains as much as Darwin's realization that species evolve over time to adapt to their environment. The way in which this happens, by natural selection of inheritable characteristics, is both so clever and so simple it begs the question why the world needed to wait until the second half of the 19th century before someone thought of it. Nevertheless, it took another century and the advent of the computer before researchers would become interested in simulating the process of natural evolution.

Although researchers in the late 1950s and early 1960s had developed what we would now label as evolutionary algorithms, their main aim was not to solve optimization problems, but to study the phenomenon of natural evolution. The insight that the principles of natural evolution could be used to solve optimization problems in general, came in the early 1960s, when researchers like \citet{box1957evolutionary}, \citet{friedman1959digital}, and several others had independently developed algorithms inspired by evolution for function optimization and machine learning. One of the first methods to receive some share of recognition was the so-called evolution strategy \citep[as later reported in][]{rechenberg:89}. Evolution strategy was still quite far from what we would call an evolutionary algorithm: it did not use a population or crossover. One solution (called the parent) was mutated and the best of the two solutions became the parent for the next round of mutation.

Evolutionary programming, introduced a few years later \citep{fogel.owens.ea:66}, represented solutions as finite-state machines, but also lacked the concepts of both a population and crossover.  The true start of the field of evolutionary algorithms came with the seminal work of John Holland \citep{holland:75}, who was the first to recognize the importance of both concepts. With his schemata-theorem, that essentially states that high-quality schemata (``parts'') of solutions will  increase in frequency in successive iterations of the algorithm, Holland was also among the pioneers of theory-building in metaheuristics. The schemata theory was criticised later for its limited use and lack of general applicability \citep[e.g.,][]{colin2002genetic}, but it demonstrated that the field of metaheuristics needed not forever be devoid of theoretical underpinning.

It was perhaps the book by \citep{goldberg:89} (a student of John Holland) that truly sparked the evolutionary revolution. Evolutionary methods became extremely popular, journals and conferences specifically devoted to this topic sprouted and an exponentially increasing number of papers appeared in the literature. A large number of variants were proposed, each with its own specific characteristics. Extraordinary claims were made, not necessarily grounded in empirical evidence. The quest for a generic heuristic optimization method that could solve any problem efficiently, without requiring problem-specific information, seemed finally to be on the right track.

In the 1980's, the first papers start to appear that introduced general problem-solving frameworks not based on natural evolution. One of the first used another metaphor: annealing, the controlled heating and cooling process used in metallurgy and glass production to remove stresses from the material \citep{kirkpatrick.gelatt.ea:83}. Simulated annealing used random solution changes and ``accepted'' these if they improved the solution or, if they did not, with a probability inversely proportional to the solution quality decrease and proportional to an external parameter called the ``temperature''. 

For a while, it might have seemed that the development of metaheuristics was all about finding a suitable process to imitate. The 80s, however, also saw the development of several methods that reached back to the early period and used ideas derived from human problem-solving. One of the most powerful ideas was that solutions could be gradually improved by iteratively making small changes, called \emph{moves}, to them. To this end, an algorithm would investigate all or some of the solutions that could be reached from the \emph{current} solution by executing a single move. Together, these solutions form the \emph{neighborhood} of the current solution. 

Threshold accepting \citep{dueck.scheuer:90}, a simple variant of simulated annealing demonstrated that a metaphor was certainly not necessary to develop a powerful general-purpose optimization framework. The great deluge method and record-to-record travel \citep{dueck:93} differed from threshold accepting only by the way in which they accepted new solutions. Still, each of these were seen as different methods. 

Perhaps the most influential of the AI-based methods was tabu search \citep{glover:86}. The basic premise of this framework that a local search algorithm could be guided towards a good solution by using some of the information gathered during the search in the past. To this end, the tabu search framework defined a number of memory structures, that captured aspects of the search. The most emblematic is without a doubt the \emph{tabu list}, a list that records attributes of solutions and prohibits, for a certain number of iterations, any solutions that exhibit an attribute on the tabu list. 

The same paper that introduced tabu search also coined the word  ``meta-heuristic'' \citep{glover:86}. However, not everybody agreed with this term and a push was made to use the (more modest) term ``modern heuristics'' instead. Clearly, not everybody agreed that the limited set of metaheuristics proposed by the 1980s had a higher-level aspect to them. Many still viewed them essentially as algorithms (admittedly, more complicated than their simple counterparts), i.e., unambiguous step-by-step sets of operations to be performed. Indeed, it is very common in the late 80s for a ``new metaheuristic'' to be described based on a flowchart or another typical algorithmic representation. The widespread realization that metaheuristics could and should be viewed as general frameworks rather than as algorithms, would come during the next period, the framework-centric period. 

Interestingly, neural networks \citep{hopfield:82} were among the limited list of metaheuristics proposed by the late 1980s. These methods imitate the functioning of a brain (including neurons and synapses) and were originally proposed in the context of pattern recognition (for which they are still mostly used). 

A development of the late 1980s that presaged the framework-centric period came from the adaptive memory orientation of tabu search, embodied in a search principle that underlies most modern metaheuristic algorithms; namely, an emphasis on incorporating strategies to achieve the goals of intensification and diversification \citep{glover1989tabu}. Intensification refers to strategies that reinforce attributes (such as assignments of values to variables) of solutions with high evaluations from among those visited in the past, while diversification refers to strategies that generate solutions containing attributes that have never (or rarely) been incorporated into past solutions. From a neighborhood orientation, intensification strategies focus on regions where high quality solutions have been found, while diversification strategies focus on driving the search into new regions that exhibit characteristics differing from those previously encountered. The intensification/diversification focus, while seeming natural and inevitable today, contrasted with perspectives that prevailed at the time (and which still exert an influence) which relied chiefly on randomization and gave little attention to the relevance of memory (and hence learning) within metaheuristics. 

An excerpt from the book by \citet{glover1997tabu} conveys the underlying theme which has now been adopted by most modern metaheuristics, and highlights the difference between this theme and the primary alternative that previously enjoyed a wide popularity which derived from control theory: 
  
{\itshape\small
The relevance of the intensification/diversification distinction is supported by the usefulness of [memory-based] strategies that embody these notions.  Although both intensification and diversification operate in the short term as well as the long term, longer term strategies are generally the ones where these notions find their greatest application.

In some instances, we may conceive of intensification as having the function of an intermediate term strategy, while diversification applies to considerations that emerge in the longer run.  This view comes from the observation that in human problem solving, once a short term strategy has exhausted its efficacy, the first (intermediate term) response is often to focus on the events where the short term approach produced the best outcomes, and to try to capitalize on elements that may be common to those events.  When this intensified focus on such events likewise begins to lose its power to uncover further improvement, more dramatic departures from a short term strategy are undertaken.  (Psychologists do not usually differentiate between intermediate and longer term memory, but the fact that memory for intensification and diversification can benefit from such differentiation suggests that there may be analogous physical or functional differences in human memory structures.)  Over the truly long term, however, intensification and diversification repeatedly come into play in ways where each depends on the other, not merely sequentially, but also simultaneously.

There has been some confusion between the terms intensification and diversification, as applied in tabu search, and the terms exploitation and exploration, as popularized in the literature of genetic algorithms.  The differences between these two sets of notions deserve to be clarified, because they have substantially different consequences for problem solving.

The exploitation/exploration distinction comes from control theory, where exploitation refers to following a particular recipe (traditionally memoryless) until it fails to be effective, and exploration then refers to instituting a series of random changes — typically via multi-armed bandit schemes — before reverting to the tactical recipe.  (The issue of exploitation versus exploration concerns how often and under what circumstances the randomized departures are launched.)

By contrast, intensification and diversification in tabu search are both processes that take place when simpler exploitation designs play out and lose their effectiveness — although as we have noted, the incorporation of memory into search causes intensification and diversification also to be manifest in varying degrees even in the short range.  (Similarly, as we have noted, intensification and diversification are not opposed notions, for the best form of each contains aspects of the other, along a spectrum of alternatives.)

Intensification and diversification are likewise different from the control theory notion of exploration.  Diversification, which is sometimes confused with exploration, is not a recourse to a Game of Chance for shaking up the options invoked, but is a collection of strategies — again taking advantage of memory — designed to move purposefully rather than randomly into uncharted territory.

The source of these differences is not hard to understand.  Researchers and practitioners in the area of search methods have had an enduring love affair with randomization, \ldots and many find a special enchantment in miraculous events where blind purposelessness creates useful order.  We have less often been disposed to notice that this way of producing order requires an extravagant use of time, and that order, once created, is considerably more effective than randomization in creating still higher order.
}

The transition to this modern perspective on intensification and diversification, and its importance in the development of effective metaheuristics, is also noted in \citet{blum2003metaheuristics}.   

By 1995, research in metaheuristics had grown to a level that could sustain its own conference series and thus the MIC (Metaheuristics International Conference) series was established. In the same year, the first issue of the Journal of Heuristics\footnote{\url{http://link.springer.com/journal/10732}}, for many years the only journal dedicated to publishing research in metaheuristics, and still the vanguard among such journals, was published. 

Several other frameworks that had been proposed around the early 90s, also gained increasing interest during the mid 90s. The innovation proposed in the GRASP (greedy randomized adaptive search procedure) framework was to modify a greedy heuristic by selecting at each iteration not necessarily the best element, but one of the best elements randomly \citep{feo.resende:95}. 
Similarly, Ant Colony Optimization \citep{colorni.dorigo.ea:92} proposed not only to mix deterministic and stochastic information, but also proposed a way for solutions to exchange information.

Due to the convergence and no free lunch results in the second half of the 1990s, a systematic rigorous study of the performance and behaviour of heuristics such as evolutionary algorithms and ant colony optimisation commenced \citep[see,][]{oliveto2007time,auger2011theory,neumann}. These studies discovered both easy problems where heuristics perform well, and similarly easy problems where they fail and require exponential optimisation time. Furthermore, they could rigorously prove how heuristics could efficiently optimize several classical combinatorial problems, and how they could deliver good approximations for NP hard problems.

In the same period, however, it gradually became clear that metaheuristics based on metaphors would not necessarily lead to good approaches. The promised black-box optimizers that would always “just work” and that had attracted so much attention, seemed elusive. The convergence results obtained for simulated annealing \citep{granville.krivanek.ea:94}, because they only worked when an infinite running time was available, were not as compelling for practical situations as initially thought. Similarly, the automatic detection of good building blocks by genetic algorithms only really worked if such building blocks actually existed, and if they were not continually being destroyed by the crossover and mutation operators operating on the solutions. Overall, the practical usefulness of the theoretical results that have been published mainly in the evolutionary algorithms community, is still widely debated \citep[see, e.g.,][]{mitchell1998introduction}. 

Even though the early metaheuristic frameworks offered some compelling ideas, they did not remove the need for an experienced heuristic designer. The advent of metaheuristics had not changed the simple fact that a metaheuristic that extensively exploited the characteristics of the optimization problem at hand would almost always be superior to one that took a black-box approach, regardless of the metaheuristic framework used.  

In general, researchers during the method-centric period proposed \emph{algorithms}, i.e., formalized structures that were meant to be followed like a cook book recipe. More often than not, the ``new metaheuristic'' was given a name, even when the difference between the new method and an existing method was small. 

\section{Period 3: The \textcolor{black}{framework-centric period}}

The insight that metaheuristics could be more usefully described as high-level algorithmic \emph{frameworks}, rather than as algorithms, was a natural thing to happen. The main indicator that this mindset change was taking place ---~a change that has given rise to a period that we have dubbed the \emph{framework-centric} period~--- is the increasing popularity of so-called ``hybrid'' metaheuristics during the early 2000s. Indeed, this period could by rights have been called they ``hybrid metaheuristic period''. Whereas earlier researchers used to restrain themselves to a single metaheuristic framework, more and more researchers around the turn of the century combined ideas from different frameworks into a single heuristic algorithm. Some combinations became more popular than others, like the use of a constructive heuristic to generate an initial solution for a local search algorithm, or the use of GRASP to generate solutions that are then combined using path relinking.

One type of \textcolor{black}{hybrid metaheuristic} even received a distinct name: the use of local search (or any ``local learning'' approach) to improve solutions that are obtained by an evolutionary algorithm was called a \emph{memetic} algorithm \citep{moscato:89}. In 2004, the term ``hybrid metaheuristic'' had become common, and a new conference series with the same name was started.

The hybridization of metaheuristics, however, did not restrict itself to a combination of a metaheuristic with another metaheuristic. Opening up the individual algorithmic frameworks allowed researchers to combine a metaheuristic with any auxiliary method available. Constraint programming, linear programming and mixed-integer programming, were all used in combinations with ideas from metaheuristics. The combination of metaheuristics and exact methods was coined ``matheuristics'' \citep{maniezzo.stutzle.ea:10} (though these methods too had many antecedents in the metaheuristics literature). In 2006, the first edition of the Matheuristics conference took place.

Soon after its introduction, the term ``hybrid'' metaheuristic would become obsolete, as researchers made a general transition from seeing metaheuristics as algorithms of which some components could be borrowed by other metaheuristics, to general sets of concepts (``frameworks''). The ``metaheuristic framework'' concept entailed that metaheuristics were nothing more (or less) than a more or less coherent set of ideas, which could of course, be freely combined with other ideas. Today, many researchers develop metaheuristics using their experience and knowledge about which methods will work well for certain problems and which most likely will not. 

Some general patterns started to appear in the literature on which methods work well for which problems, and the community gravitated towards approaches that always delivered. For almost any variant of the vehicle routing problem, e.g., a large majority of approaches use some form of local search as their main engine, generally in a multi-neighborhood framework like variable neighborhood search \citep{mladenovic.hansen:97}. The use of several different local search operators or the use of several different constructive procedures in general is now a well-regarded strategy and often used as the first choice by heuristic designers. Clearly, variable neighborhood search presented a framework within which the use of multiple neighborhoods could be captured, but many other ways of combining several local search operators in a single heuristic are possible. 

Crucial in this period, which is still ongoing, is that researchers do not have to propose a ``new algorithm'' anymore to get their papers published. By combining the most efficient operators of existing metaheuristic frameworks, and carefully tuning the resulting heuristic, algorithms can be created that solve any real-life optimization problem efficiently. Researchers can now focus on studying a single, mundane aspect of a metaheuristic framework in detail like, e.g., its stopping rules \citep{ribeiro.rosseti.ea:11}. 

Even though the scientific community has made considerable progress in its quest to understand the fundamental behavior of metaheuristics (See Section~\ref{sec:method}), the metaheuristic community has traditionally put a heavy focus on \emph{performance}. Research is only considered good if (and only if) it produces a heuristic algorithm that ``performs'' well with respect to some benchmark, such as another heuristic or a lower bound. This has been called the ``up-the-wall game'' (though it might also be called the ``one-upmanship game''). All other contributions (e.g., heuristics that are many times simpler than the best-performing heuristic in the literature, studies on heuristics that should perform well but for some reason do not, \ldots) are much more difficult to publish. However, several researchers have pointed out the adverse effects of this paradigm (which effectively reduces science to a game), and some recent contributions that go beyond the up-the-wall game demonstrate the framework-centric period is gradually transforming into the \emph{scientific} period. In this period the study of metaheuristics will shift its focus from performance to understanding. Unfortunately, however, not all of the metaheuristics community makes the transition to the framework-centric period, and we are forced to report on a period which essentially runs in parallel with this  period. 

\section{The \textcolor{black}{metaphor-centric period}}

Starting in the 1980s a sub-field has arisen of research (we hesitate to put quote marks around the term for reasons that will be explained below) that focuses on the development of new metaheuristic methods based on metaphors of natural or man-made processes. In our history, this period has not been assigned a number because it does not fit chronologically between the other periods, but rather is a sidestep that happened (and is still happening) in parallel to the framework-centric period. 

Although metaphors had been useful in the development of early metaheuristics as a source of inspiration for the development of novel frameworks, it has always been evident to many that a metaphor is only a metaphor and always breaks down at a certain point. It is therefore useful for inspiration, but not necessarily everything about it usefully translates to a metaheuristic framework. Importantly, a metaphor is not enough to \emph{justify} metaheuristic design choices or to create a foundation for completely new metaheuristics.

In recent years, however, a different attitude seems to have taken hold of a subfield of the metaheuristics community. The aim of the ``metaphor-based'' subfield  seems to center entirely around the development of ``novel'' metaphors that can be used to motivate new metaheuristics. The list of natural and man-made processes that have inspired such metaheuristic frameworks is huge. Ants, bees, termites, bacteria, invasive weed, bats, flies, fireflies, fireworks, mine blasts, frogs, wolves, cats, cuckoos, consultants, fish, glowworms, krill, monkeys, anarchic societies, imperialist societies, league championships, clouds, dolphins, Egyptian vultures, green herons, flower pollination, roach infestations, water waves, optics, black holes, the Lorentz transformation, lightning, electromagnetism, gravity, music making, ``intelligent'' water drops, river formation, and many, many more, have been used as the basis of a ``novel'' metaheuristic technique.

Moreover, there does not seem to be any restriction on the type of process that can be translated into a metaheuristic framework. One would expect that, at the very least, the process that is to become the basis for a metaheuristic should \emph{optimize} something (e.g., an annealing process \emph{minimizes} the energy level, natural evolution \emph{minimizes} the discrepancy between the characteristics of a species and the requirements of this species' environment, ants \emph{minimize} the distance between their nest and their source of food). Nevertheless, many metaheuristic frameworks can now be found based on processes that by no stretch of imagination can be said to optimize anything, like fireworks, mine blasts, or cloud formation.

Both the causes and the consequences of this ``metaphor fallacy'' have been extensively dealt with in a number of other publications \citep{sorensen:15,weyland:10} (short summary: it is not science) and this is not the place to repeat all the arguments why metaphor-based metaheuristics are a bad idea. Nevertheless, metaphor-based ``novel'' metaheuristics take up a (dark) page in the history of metaheuristics. A page that should be turned quickly.

\section{Recent innovations and applications}

Recent innovations in metaheuristics can be usefully classified under the headings of Exploiting Conditional Relationships, Attribute Selection Strategies, Unified Models and Metaheuristics with Learning.  

\subsection{Exploiting Conditional Relationships}

The significance of recognizing and exploiting conditional relationships has recently become a theme for developing more effective metaheuristics. For example, principles for exploiting such relationships have recently been elaborated to produce a class of methods called Multi-Wave Algorithms \citep{glover:16a}, whose preliminary implementation has achieved notable success \citep{oualid:17}. These principles, which are applicable both to multi-start constructive methods and iterated local search methods, may be briefly summarized as follows. 

Define a boundary solution to be a local optimum for a neighborhood search method or a completed construction for a constructive solution method, and define a solution wave to be a succession of moves, starting from a given initial solution (which can be a null solution for a constructive method) that lead to a boundary solution.  Attention is given to forms of constructive or neighborhood search that permit moves to be dropped (or equivalently, reversed) en route from an initial solution to a boundary solution. Hence, constructive search is treated within the framework of strategic oscillation, which permits destructive moves at intermediate stages and neighborhood search is treated in a similar manner, where move reversals are employed at specific junctures. The following observations motivate the use of such a framework.

\emph{Principle of Marginal Conditional Validity (MCV Principle)}.  Starting from a given initial solution, as more moves are made, the information that permits these moves to be evaluated becomes increasingly effective for guiding the search to a high quality boundary solution, conditional upon the decisions previously made. 

This principle has several consequences, which can be stated as heuristic inferences.

\emph{Inference 1}.  Early decisions are more likely to be bad ones.

\emph{Inference 2}.  Early decisions are likely to look better than they should (i.e., receive higher evaluations than would be appropriate for achieving global optimality), once later decisions have been made.

\emph{Inference 3}.  The outcome of a constructive or local improvement method can often be improved by examining the resulting complete solution, where all decisions have been made, and seeing whether one of the decisions can now be advantageously replaced with a different one.

\emph{Inference 4}.  The outcome of a constructive or improvement method can often be improved by examining a solution at an intermediate stage, before reaching a boundary solution, and seeing whether one of the decisions can now be advantageously replaced with a different one.

These conditionality principles are reinforced by an associated set of principles based on the notion of persistent attractiveness, which is manifested in three types of situations. 

\emph{PA Type 1}: Exhibited by a move that is attractive enough more than once during a wave to be chosen a second or third time (a move that receives a high enough evaluation after being dropped to cause it to be chosen again).

\emph{PA Type 2}: Exhibited by a move that repeatedly has high evaluations during a wave, whether chosen or not, and if ultimately chosen is not selected until a step that occurs somewhat after it first appears to be attractive.

\emph{PA Type 3}: Exhibited by a move that repeatedly has a high evaluation during one or more waves, but which is not selected. 

Principle of Persistent Attractiveness. 
\begin{description}
  \item[(PA-1)] A move of PA Type 1 allows a focus on dropping moves that will have greater impact by exempting such a move from being dropped anew once it achieves the Type 1 status. 
  \item[(PA-2)] A move of PA Type 2 which was chosen on a given wave offers an opportunity for producing a better solution on a new wave if it is selected at an earlier stage in the wave. (In particular, given that the move was eventually chosen, selecting it earlier affords a chance to make decisions based on this choice that were not influenced at the later point in the decision sequence when the move was chosen previously). 
  \item[(PA-3)] A move of PA Type 3 can be used to combine diversification with intensification by choosing at least one member of this type at a relatively early stage of a wave, assuring that the resulting solution will not duplicate a previous solution and allowing subsequent decisions to take this move into account.
\end{description}

These types of conditional relationships are amplified in the design of multi-wave algorithms by reference to an associated principle for combining attractiveness and move influence and are conjectured to be fundamental to designing more effective uses of memory in metaheuristic search. 

\subsection{Attribute Selection Strategies}

The genetic algorithm literature points out the relevance of encoding solutions properly in order to allow crossover operations to generate offspring that can be meaningfully exploited. The flip side of this for adaptive memory algorithms is to select particular attributes of solutions for defining tabu status (i.e., for determining conditions under which certain moves should be prohibited) in order to produce solutions of highest quality. Recent studies by \citet{wang.wu.ea:17} and by \citet{wu.wang.ea:17} have documented the impact of attribute selection strategies by employing a balance between attributes consisting of hash coded solutions (drawing on ideas of \citet{woodruff.zemel:93}) and attributes more commonly used. Computational experiments show that such blends of attributes yield algorithms that significantly outperform all other methods previously proposed for the classes of problems tested. These outcomes motivate the exploration of other ways of generating compound solution attributes to identify attributes that likewise lead to more effective metaheuristics.

\subsection{Unified Models and Applications}

A wide range of metaheuristic applications have shown to be encompassed in the quadratic unconstrained binary optimization (QUBO) model. By means of special reformulation techniques, these include many constrained optimization models as well. For example, \citet{kochenberger.hao.ea:14} identify important QUBO applications in the areas of quadratic assignment, capital budgeting, task allocation, distributed computer systems, maximum diversity, symmetric and asymmetric assignment, constraint satisfaction, set partitioning, warehouse location, maximum clique and independent sets, graph coloring, graph partitioning, number partitioning and linear ordering, among a variety of others.

A highly effective metaheuristic for solving QUBO problems identified in \citet{wang.ni.ea:15} has led to additional applications in the field of quantum computing, as documented in \citet{rosenberg.vazifeh.ea:16} and \citet{mniszewski.negre.ea:16}, culminating in an open source hybrid quantum solver called Qbsolve released by D-Wave Systems \citep{booth.reinhardt.ea:17} and reviewed in Wired magazine \citep{finley:17}.  These developments promise to bring metaheuristics into the forefront of the burgeoning field of quantum computing. 

\subsection{Metaheuristics with Learning}

Adaptive memory and strategies for exploiting it, as originally introduced in tabu search, have become incorporated into a variety of other metaheuristics with the recognition that learning (and hence memory) is essential to developing better methods. A novel approach that pursues this theme by attempting to emulate human learning in a metaheuristic setting has been undertaken in \citet{wang.ni.ea:15}. In addition, an important initiative has been launched by the LION Laboratory at the University of Trento that joins memory-based learning with metaheuristic optimization \citep{battiti.brunato:14,battiti:16}.

Proposals for joining adaptive memory with classical forms of learning go back a long way. For example, the paper that lays the foundation for scatter search and strategic oscillation \citep{glover:77} points out the relevance of using clustering in connection with strongly determined and consistent variables, whose identification depends on frequency-based memory applied to high quality solutions. In general, within a memory-based method, clustering can take advantage of the fact that solutions generated over time may be exploited more effectively by analyzing subsets of good solutions that exhibit common characteristics. Path relinking provides an indirect means for accomplishing this, as observed in connection with the successful learning approach in \citet{lai.hao.ea:16}. 

Classification, another form of classical learning, can be exploited in retrospective analysis, particularly in relation to a learning strategy called target analysis, to identify characteristics of moves that lead to elite solutions, as noted in \citet{glover.greenberg:89} and \citet{glover1997tabu}. In reverse, adaptive memory search methods have been utilized as a means to obtain improved learning procedures, as in neural networks \citep{kelly.rangaswamy.ea:96,dengiz.alabas-uslu.ea:08} and in clustering \citep{cao.glover.ea:15}. Nevertheless, the exploration of classical forms of learning within metaheuristics remains a topic that has not been covered in nearly the depth that it deserves. Key observations related to this topic emerge in the context of clustering and intensification/diversification strategies, where spatially defined measures of distance may be inapplicable \citep{glover:16b}.  

A significant recent emphasis on learning has been launched under the banner of “Oppo\-sition-based learning (OBL)” introduced in \citet{tizhoosh:05}. OBL has become the focus of numerous research initiatives in machine learning and metaheuristic optimization, and a variety of proposals and studies have undertaken to exploit its underlying ideas, as documented in the survey of \citet{xu.wang.ea:14}. Still more recently, metaheuristic intensification and diversification strategies have been discovered to provide a more flexible and comprehensive learning framework called Diversification-Based Learning \citep{glover.hao:17}, which makes it possible to remedy two principal limitations of Opposition-Based Learning, by which OBL fails to give a direct means for establishing feasibility and is compelled to resort to randomization when its framework breaks down. Once again, this work shows the value of giving increased attention to the topic of learning in future metaheuristic studies.   

\subsection{Recent Metaheuristic Applications in Practice}

It is impossible to cover even a fraction of the practical applications of metaheuristics, but it may be useful to point out a few salient applications that have recently emerged to give a sense of where the field is making an impact. We have already noted a variety of these applications under the heading of ``Unified Models and Applications''. New developments abound for solving classical graph theory and related problems, which are beyond the scope of this article to catalog. Here we undertake to round out the picture for practical applications apart from these more academic problems by indicating a few more that bear mention. Table~\ref{tab:pbdom} reports some of these practical applications.

\begin{table}[htbp]
  \centering
  \begin{tabular}{ll}
    \hline
    \multicolumn{1}{c}{Problem domain} & \multicolumn{1}{c}{Reference} \\ \hline
    Supply chain scheduling            & \citet{pei.liu.ea:14}        \\
    Supply chain disruption mitigation     & \citet{schmitt.kumar.ea:17}\\
    Network design connectivity            & \citet{shangin.pardalos:15}\\
    Scheduling of wireless sensor networks & \citet{lersteau.rossi.ea:16}\\
    Maintenance problems                   & \citet{todosijevic.benmansour.ea:16}\\
    Dynamic memory allocation problems     & \citet{sanchez-oro.sevaux.ea:15} \\
    Multi-depot vehicle routing            & \citet{li.pardalos.ea:15}\\
    Clustered capacitated vehicle routing  & \citet{exposito-izquierdo.rossi.ea:16} \\
    Swap-body vehicle routing              & \citet{todosijevic.hanafi.ea:17} \\
    Water distribution network design      & \citet{corte.sorensen:16} \\
    Critical node problems                 & \citet{aringhieri.grosso.ea:16}\\
    Generalized cable-trench problems      & \citet{vasko.landquist.ea:15}\\
    Multi-depot ring star problems         & \citet{hill.voss:16}\\
    Job scheduling with smoothing costs    & \citet{respen.zufferey.ea:15}\\
    Static data segment location           & \citet{sen.krishnamoorthy.ea:16}\\
    Wireless multi-role sensor networks    & \citet{castano.bourreau.ea:16}\\
    Hub location routing                   & \citet{lopes.andrade.ea:16}\\
 \hline   
  \end{tabular}
  \caption{Problem domains and key references}
  \label{tab:pbdom}
\end{table}

\section{Period 4: The \textcolor{black}{scientific period}?}

For a long time, the field of metaheuristics has had difficulties to be taken seriously. In 1977, one of the authors of this chapter wrote ``[exact] algorithms are conceived in analytic purity in the high citadels of academic research, heuristics are midwifed by expediency in the dark corners of the practitioner's lair [\dots] and are accorded lower status.'' \citep{glover:77} Traditionally, the theoretical underpinning of heuristics and metaheuristics has not been on par with that of other areas in OR, more specifically exact methods. The development of heuristic optimization algorithms, whether using a metaheuristic framework or not, is guided by experience, not theory. Early attempts to firmly ground the development of metaheuristics in theory have not delivered upon their promises.  Understanding the behavior of metaheuristics on a fundamental level has proven to be a difficult task, notwithstanding several noteworthy efforts \citep[e.g.,][]{watson.barbulescu.ea:02,watson.beck.ea:03}. 

Nevertheless, it is hard to argue with success. The obvious usefulness of metaheuristics in practical optimization problems has drawn researchers to improve the frameworks and methods developed. To solve a large majority of real-life optimization problems, heuristics are and will remain the only option, whether developed using a metaheuristic framework or not. Nevertheless, there are many things that can be improved about the way the metaheuristic community operates. To list just a few:
\begin{itemize}
    \item The establishing of adequate testing protocols, to ensure that algorithms perform as well as they are claimed to do. 
    \item The introduction of meta-analysis (i.e., a review of a clearly formulated question that uses systematic methods to identify, select, and evaluate relevant research, as well as to collect and analyze data from relevant studies) to the field of metaheuristics \citep{hvattum-meta}.
    \item The requirement to disclose source code, so that researchers can check and build on each other's work without in a more efficient way, without reinventing the wheel. 
    \item The development of powerful general-purpose heuristic solvers to decrease development time, as by starting from a foundation of a powerful exact solver like CPLEX or Gurobi, and then modifying it to operate heuristically. LocalSolver seems to be on the right track.
    \item Supporting these general-purpose solvers, the development of a powerful and generally accepted modeling language, geared more towards the development of heuristics and less towards the MIP-paradigm.
    \item \ldots
\end{itemize}

Most importantly, the change from a performance-driven community to a community in which scientific understanding is more important, will take place during the scientific period. Without doubt, this will lead to the development of even better heuristics, even more efficient, but it will also lead to heuristics that are usable outside of the developer's lab environment.  

\section{Conclusions}

Describing the history of the field of metaheuristics in a few pages is not an easy undertaking and completeness is a goal that simply cannot be achieved. In this chapter, we have attempted to clarify the evolution in this field by not focusing exclusively on important events or publications, but by attempting to identify the important paradigm shifts that the field has dealt with. What is certain, is that the use of metaheuristics is older, much older, than the term itself. As mentioned, our brain itself houses some powerful metaheuristics that have helped humans survive from the dawn of mankind. The scientific study of metaheuristics, however, had to wait until the second half of the previous century.

Scientific communities invariably develop a conceptual framework within which a few axioms are held to be true. This can also be said of the metaheuristics community. It is those shared truths that we have attempted to uncover in this chapter. Even though the field of metaheuristics is still young, it has already undergone several paradigm shifts that have changed the way researchers look upon the development of heuristic optimization methods.

The transition from the method-centric to the framework-centric period has been beneficial for the entire community, and there is no doubt that the transition towards the scientific period can take the field further into the right direction. Metaheuristics are a fascinating area of study with highly significant practical ramifications and the field will certainly keep on evolving in the foreseeable future. There is no doubt that a more scientific, less dogmatic, and broader point of view can help us all in achieving our goals: the  development of efficient methods to solve the most challenging and important real-life optimization problems.

\bibliographystyle{plainnat}

\end{document}